\documentclass[11pt]{article}
\usepackage[a4paper]{geometry} 
\usepackage{clic2021} 
\usepackage{times} 
\usepackage{xurl} 
\usepackage[italian,english]{babel}
\usepackage{latexsym} 
\usepackage{xcolor}
\usepackage{amsmath}
\usepackage{soul}
\usepackage[utf8x]{inputenc}

\newcommand\blfootnote[1]{%
  \begingroup
  \renewcommand\thefootnote{}\footnote{#1}%
  \addtocounter{footnote}{-1}%
  \endgroup
}

\newcommand{\sara}[1]{{\textcolor{
black}{#1}}}

\title{Visualization: the missing factor in Simultaneous Speech Translation}

\author{\textbf{Sara Papi$^{1,2}$, Matteo Negri$^{1}$, Marco Turchi$^{1}$} \\
  1. Fondazione Bruno Kessler, Italy \\
  2. University of Trento, Italy \\
  \tt{\{spapi,negri,turchi\}@fbk.eu}}

\date{}

\begin{document}
\maketitle
\begin{abstract}
Simultaneous speech translation (SimulST) is the task 
in which 
output generation has to be performed on partial, incremental speech input.
In recent years, SimulST has become popular due to the spread of multilingual application scenarios, like international live conferences and streaming lectures, in which on-the-fly speech translation 
can facilitate users' access to audio-visual content. In this paper, we analyze the characteristics of the SimulST systems developed so far,
discussing 
their strengths and weaknesses.
We then concentrate on the evaluation framework required to 
properly assess systems' effectiveness.
To this end, we raise the need for a broader performance analysis, also including the user experience standpoint. We argue that SimulST systems, indeed, 
should
be evaluated not only in terms of quality/latency measures, but also via task-oriented metrics accounting, for instance, for the visualization strategy adopted.
In light of this, we 
highlight which are the goals achieved by the 
community
and what is still missing.\blfootnote{Copyright \copyright 2021 for this paper by its authors. Use permitted under Creative Commons License Attribution 4.0 International (CC BY 4.0).}
\end{abstract}

\section{Introduction}

%
Simultaneous speech translation (SimulST) is the task in which the translation of a source language speech 
has to be performed on partial, incremental input.
This is
a key feature to achieve low latency in scenarios like streaming conferences and lectures, where the text has to be displayed following as much as possible the pace of the speech.
SimulST is indeed a 
complex task in which the difficulties of performing speech recognition from partial inputs are exacerbated by the problem to project meaning across languages.
Despite the increasing demand for such a system, the problem is still far from being solved.
%
%
%

      

So far, research efforts mainly focused on the quality/latency trade-off, i.e. producing high quality outputs in the shortest possible time, balancing the need for a good translation with the necessity of a rapid text generation. 
Previous
studies, however, disregard how the translation is displayed
and, consequently, how it is actually 
perceived by the end users. 
After a concise survey of the state of the art in the field, in this paper we posit  
%
that, from the users' experience standpoint, output visualization is at least as important as having a good translation in a short time.
This raises the need for a broader, task-oriented and human-centered analysis of 
SimulST systems' performance, also accounting for 
this third crucial factor.
%
%

\section{Background}
As in the case of offline speech translation, 
the adoption of cascade architectures \cite{StentifordSteer88,Waibel1991b} was the first attempt made by the 
SimulST community to tackle the problem of generating text 
from partial, incremental input.
Cascade systems
\cite{Fgen2009ASF,Fujita2013SimpleLC,Niehues2018LowLatencyNS,xiong2019dutongchuan,arivazhagan2020re} 
involve a pipeline of two components.
First, a streaming automatic speech recognition (ASR) module transcribes the input speech into the corresponding text \cite{wang2020low,moritz2020streaming}. Then, a simultaneous text-to-text translation module translates the partial transcription into 
target-language text \cite{gu-etal-2017-learning,dalvi-etal-2018-incremental,ma-etal-2019-stacl,arivazhagan-etal-2019-monotonic}.
This
approach
suffers from \textit{error propagation}, a well-known problem even in the offline scenario, where the transcription errors made by the ASR module are propagated to the MT module, which cannot recover from them
as it does not have direct access to the audio.
Another strong limitation of 
cascaded systems
is the \textit{extra latency} added by the two-step pipeline, since the MT module has to wait until the streaming ASR output is produced.

To overcome these issues, the direct models initially proposed in \newcite{berard_2016,weiss2017sequence} represent a valid alternative that is gaining increasing traction \cite{bentivogli-etal-2021-cascade}.
Direct ST models are composed of an encoder, usually 
bidirectional,
and a decoder. The encoder starts from the audio features extracted from the input signal and computes a hidden 
representation; the
decoder transforms this representation into 
target language text.
Direct modeling becomes 
crucial in the simultaneous scenario, as it reduces the overall system's latency 
due to the absence of intermediate 
symbolic representation steps.
Despite the data scarcity issue caused by the limited availability of speech-to-translation corpora,
the adoption of direct architectures showed to be promising 
\cite{weiss2017sequence,ren-etal-2020-simulspeech,zeng-etal-2021-realtrans}, driving recent efforts towards the development of increasingly powerful and efficient models.

\section{Architectural Challenges} 

This section surveys the direct SimulST models developed so far, highlighting strengths and weaknesses of the current architectures and decision policies -- i.e. the strategies used by the system to decide whether
to output a partial translation or to wait for more audio information. 
We discuss ongoing research on architectural improvements of encoder-decoder models, as well as popular approaches like offline training and re-translation.
All these works concentrate on reducing systems latency, targeting a better quality/latency 
trade-off.

\paragraph{Encoding Strategy.}
Few
studies \cite{Elbayad2020,nguyen21d_interspeech} 
tried
to improve 
the
encoder part of simultaneous systems.
\newcite{Elbayad2020} and \newcite{nguyen21d_interspeech} introduced the use of unidirectional encoders instead of standard  bidirectional encoders (i.e. the encoder states are not updated after each read action) to speed up
the decoding phase. \newcite{nguyen21d_interspeech} also proposed an encoding strategy called \textit{Overlap-and-Compensate}, where the encoder exploits extra frames provided from the past that were discarded during the previous encoding step. 
The segmentation problem is a crucial aspect in SimulST, where the system needs to split a long audio input into smaller chunks (speech frames) in order to process them. Different segmentation techniques can be adopted to extract this information, starting from the easiest one based on fixed time windows \cite{ma-etal-2020-simulmt} to the dynamic ones based on automatically detected word boundaries \cite{zeng-etal-2021-realtrans,chen-etal-2021-direct}. \newcite{ma-etal-2020-simulmt} also studied the dynamic segmentation based on oracle boundaries but they discovered that, in their scenario, it had worse performance compared to that of the fixed segmentation.

%


\paragraph{Decoding Strategy.}
%
Some efforts have been made
to improve the decoding strategy as it strongly correlates to the decision policy  of 
simultaneous
systems.
Speculative beam search, or SBS, \cite{zheng-etal-2019-speculative} represents the first successful attempt to use beam search in SimulST.
This technique consists in hallucinating several prediction steps in the future in order to make more accurate decisions based on the best ``speculative" prediction obtained.
Also \newcite{zeng-etal-2021-realtrans} integrate the beam search in the decoding strategy, 
developing the wait-k-stride-N strategy. 
In particular, the authors bypass output speculation by directly applying beam search, after waiting for k words, on a word stride of size N (i.e., on N words at a time) instead of one single word as prescribed by the standard wait-k. 
\newcite{nguyen2021empirical} analyzed several decoding strategies relying on different output token granularities, such as characters and Byte Pair Encoding (BPE), showing that the latter yields lower latency.

\paragraph{Offline or Online training?}
An alternative approach to simultaneous training is the offline (or full-sentence) training of the system and its subsequent use as a simultaneous one. 
\newcite{nguyen2021empirical} explored this solution with an LSTM-based direct ST system, analyzing the effectiveness of different decoding
strategies.
Interestingly, the offline approach does not only preserve overall performance despite the switch of modality, it also improves system's ability to generate well-formed sentences.
These results are 
confirmed
by \newcite{chen-etal-2021-direct}, who successfully exploit a direct ST system jointly trained in an offline fashion with an ASR one.  

\paragraph{Another point of view: re-translation.}
Re-translation \cite{Niehues2016DynamicTF,Niehues2018LowLatencyNS,arivazhagan-etal-2020-translation,arivazhagan2020re} consists in 
re-generating the output from scratch (e.g. after a fixed amount of time) for as long as new information is received.
%
%
%
This approach ensures high quality (the final output is produced with all the available context) and low latency (partial translations can be generated with fixed, controllable delay).
%
%
%
This, however, comes at the cost of 
strong output instability 
(the so-called \textit{flickering}, due to continuous updates of the displayed translations)
which is not optimal from the user experience standpoint.
To this end, some metrics have been developed to measure the instability phenomenon, such as the \textit{Erasure} \cite{arivazhagan2020re}, which measures the number of tokens that were deleted from the emitted translation to produce the next translation. 

\paragraph{Decision Policy.}
In simultaneous settings, 
the model has to decide, at each time step, if the available information is enough to produce a partial translation -- i.e. to perform a \textit{write} action using the information received until that step (audio chunk/s in case of SimulST or token/s in case of 
simultaneous MT)
-- or if it has to wait and perform a \textit{read} action to receive new information from the input.
Possible decision policies result in different ways to balance the quality/latency trade-off. On one side, more read actions provide the system with larger context useful to generate translations of higher quality.
On the other side, this counterbalances the increased, sometimes unacceptable latency. To address this problem, two types of policy have been proposed so far: fixed and adaptive.
%
While \textit{fixed} decision policies look at the  number of ingested tokens (or speech chunks, in the speech scenario), in the \textit{adaptive} ones the decision is taken by also looking at the contextual information extracted from the input.

While little research 
focused
on adaptive policies \cite{gu-etal-2017-learning,zheng-etal-2019-simpler,Zheng2020SimultaneousTP} due to the hard and time-consuming training \cite{zheng-etal-2019-simultaneous,arivazhagan-etal-2019-monotonic}, the adoption of very easy-to-train fixed policies 
is the typical choice.
Indeed, the most widely used policy is a fixed one, called 
\textit{wait-k} \cite{ma-etal-2019-stacl}.
Simple yet effective, it is based on waiting for $k$ source words before starting to generate the target sentence, as shown in Table~\ref{tab:waitk}.

\begin{table}[ht!]
\small
    \begin{tabular}{l|ccccccc}
        \textbf{source} & It & was & a & way & that & parents & ... 
        \\
        \hline
        \textbf{wait-3} & - & - & - & Es & ging & um & eine 
        \\
        \hline
        \textbf{wait-5} & - & - & - & - & - & Es & ging  
    \end{tabular}
    \caption{wait-k policy example with $k=\{3,5\}$}
    \label{tab:waitk}
\end{table}

As the original wait-k implementation is based on textual source data, \newcite{ma-etal-2020-simulmt} adapted it to the audio domain by waiting for $k$ fixed 
time frames
(audio chunks or speech frames) rather than $k$ words.
However, this simplistic approach does not consider various aspects of human 
speech,
such as different speech rates, duration, pauses, and silences.
In \cite{ren-etal-2020-simulspeech}, the adaptation was done
differently, by including
a Connectionist Temporal Classification (CTC)-based \cite{10.1145/1143844.1143891} segmentation module that is able to determine word 
boundaries.
In this case, the wait-k strategy is applied by waiting for $k$ pauses between words that are automatically detected by the segmenter.
Similarly, \newcite{zeng-etal-2021-realtrans} employed the CTC-based segmentation method but applying a \textit{wait-k-stride-N} policy to allow 
re-ranking during the decoding phase.
The \textit{wait-k-stride-N} model emits more than one word at a time,
slightly increasing the latency, since the output is prompted after the stride is processed.
This small increase in latency, however, allows the model to perform beam search on the stride, which has been shown to be 
effective in improving translation quality \cite{NIPS2014_a14ac55a}. 
Decoding more than one word at a time is the approach also employed by \newcite{nguyen2021empirical}, who showed that emitting two words increases the quality of the translation without any relevant impact on latency.
Another way of applying the wait-k strategy was 
proposed
by \newcite{chen-etal-2021-direct}, 
where a streaming ASR system is used to guide the direct ST decoding.
They look at the ASR beam to decide how many tokens have been emitted within the partial audio segment, hence having the information to apply the original wait-k policy in a straightforward way.
An interesting 
solution
is also the one 
by \newcite{Elbayad2020},
who 
jointly train a direct model across multiple wait-k paths. 
Once
the sentence has been encoded, they optimize the system by uniformly sampling the $k$ value for the decoding step. Even though they reach good performance by using a single-path training with 
$k$=7 and a different $k$ value for testing, the multi-path approach proved to be effective. 
One of its advantages is that no $k$ value has to be specified for the training, which allows to avoid the training from scratch of several models for different values of $k$.

\paragraph{Retrospective.}
\sara{All the aspects analyzed in this section highlight several research directions already taken by the simultaneous community, which have to be studied more in depth. Among all, the audio or text segmentation strategy clearly emerges as a fundamental factor of simultaneous systems, and the ambivalent results obtained in several studies point out that this aspect has to be better clarified. 
Moreover, the presence of extensive literature on the wait-k policy shows that it represents one of the topics of greatest interest to the community, which continues to work on it to further improve its effectiveness as it directly impacts on the systems' performance, especially latency.
Unfortunately, all these studies focus on the architecture enhancements and decision policies 
despite the absence of a
unique and clear evaluation framework to perform a correct and complete analysis of the system.
}

\section{Evaluation Challenges}

A good simultaneous model should produce a high quality 
translation with reasonable timing, as waiting too long will negatively affect the user experience.
Offline MT and ST communities commonly use the well-established BLEU metric \cite{papineni-etal-2002-bleu,post-2018-call} to measure the quality of the output translation, but a simultaneous system also needs a metric that accounts for the time spent by the system to output the partial translation.
Simultaneous MT (SimulMT) is the task in which a real-time translation is produced having a partial source text at disposal.
Since SimulMT was the first yet easiest simultaneous scenario studied by the community, a set of metrics was previously introduced for the textual input-output translation part.

\paragraph{Latency Metrics for SimulMT.} 
The first metric, the \textit{Average Proportion} (AP), was proposed by \newcite{cho2016neural} and measures the average proportion of source input read when generating a target prediction\sara{, that is the sum of the tokens read when generating the partial target.}
However, AP is not length-invariant, i.e. the value of the metric depends on the
input and output lengths and is not evenly distributed on the [0, 1] interval
\cite{ma-etal-2019-stacl}, making this metric strongly unreliable.

To overcome all these problems, \newcite{ma-etal-2019-stacl} introduced the \textit{Average Lagging} (AL) that 
directly describes the lagging behind the ideal policy, i.e. a policy that produces the output exactly at the same time as the speech source.
As a downside, Average Lagging is not differentiable, which is, instead, a useful property, especially if the metric is likely to be added in the system's loss computation. For this reason, \newcite{cherry2019thinking} proposed the \textit{Differential Average Lagging} (DAL), introducing a minimum delay 
after each operation. 

Another way of measuring the lagging is to compute the alignment difficulty of a source-target pair. Hence, \newcite{elbayad-etal-2020-online} proposed the \textit{Lagging Difficulty} (LD) metric that exploits the use of the \texttt{fast-align} \cite{dyer-etal-2013-simple} tool to estimate the source 
and target 
alignments. Then, they infer the reference decoding path 
and compute the AL metric.
The authors claimed the LD to be a realistic measure of the simultaneous translation as it also evaluates how a translation is easy to align considering the context available when decoding.

\paragraph{Latency Metrics for SimulST.}
The most popular AP, AL and DAL metrics were successively adapted by the SimulST community to the speech scenario by converting, for instance, the number of words
to the sum of the speech segment durations, 
as per \cite{ma-etal-2020-simuleval}.
Later, \newcite{ma-etal-2020-simulmt} raised the issue of using computational unaware metrics and proposed computational aware metrics accounting for the 
time spent by the model to generate the output.
Unfortunately, computing such metrics is not easy at all in absence of a unique and reproducible environment that can be used to evaluate the model's performance. To this end, \newcite{ma-etal-2020-simuleval} proposed \textit{SimulEval}a tool which computes the metrics by simulating a real-time scenario with a server-client scheme. This toolkit automatically evaluates simultaneous translations (both text and speech) given a customizable agent that can be defined by the user and that will depend on the adopted policy.
Despite the progress in the metrics for evaluating quality and latency, no studies have been conducted on the effective correlation with user experience.
This represents a missing key point in the current evaluation framework landscape, giving rise to the need for a tool 
that combines quality and latency metrics with application-oriented metrics (e.g., reading speed), which are strongly correlated to the visualization and, as an ultimate goal, to the user experience.

\section{The missing factor: Visualization}
In the previous section, we introduced the most popular metrics used to evaluate the simultaneous systems' performance. These metrics account for the quality and the latency of the system without capturing the user needs.
Although many researchers acknowledge the importance of human evaluation, this current partial view
can push the community in the wrong direction, in which all the efforts are focused on the quality/latency factors while the problem experienced by the user is of another kind. 
Indeed, the third factor that matters and strongly influences the human understanding of a -- even very good -- translation is the \textit{visualization strategy} adopted. The visualization problem and the need to present the text in a readable fashion for the user was only faced in our previous work \cite{karakanta-etal-2021-simultaneous}. In the paper, we raised the need for a clearer and less distracting visualization of the SimulST system's generated texts by presenting them as subtitles (text segmented in lines preserving coherent information). We proposed different visualization strategies to better assess the online display problem, attempting to simulate a setting where human understanding is at the core of our analysis.
\paragraph{Visualization modalities. }
The standard \textit{word-for-word} visualization method \cite{ma-etal-2019-stacl}, in which the words appear sequentially on the screen as they are generated, could be strongly sub-optimal for the human understanding \cite{romero-fresco-2011-respeaking}.
Infact, the word-for-word approach has two main problems: \emph{i)} the emission rate of words (some go too fast, some too slow) is irregular and the users waste more time reading the text because their eyes have to make more movements, and \emph{ii)} emission of pieces of text that do not correspond to linguistic units/chunks, requiring more cognitive effort. 
Moreover, when the maximum length of the subtitle (that depends on the dimensions of the screen) is reached, the subtitle disappears without giving the user enough time to read the last words emitted.
\sara{As this will negatively impact the user experience, we propose in \cite{karakanta-etal-2021-simultaneous} to adopt different visualization modes that better accommodate the human reading requirements.}
We first introduced the \textit{block} visualization mode, for which an entire subtitle is displayed at once (usually one or two lines maximum) as soon as the system has finished generating it. This display mode is the easiest to read for the user because it prevents re-reading phenomena \cite{Rajendran-et-al-2013} and unnecessary/excessive eye fixations \cite{Romero-fresco-2010-Standingonquicksand}, reducing the human effort. However, we discovered that the latency introduced by waiting for an entire subtitle is too high to let this visualization mode be used in many simultaneous scenarios. As a consequence, we proposed the 
\textit{scrolling lines} visualization mode that displays the subtitles line by line. 
Every time a new line becomes available, it appears at the bottom of the screen, while the previous (older) line is scrolled to the upper line. In this way, there are always two lines displayed on the screen.
To evaluate the performance of the system in the different visualization modes, we also proposed an ad-hoc calculation of the \textit{reading speed} (characters per second or CPS) that correlates with the human judgment of the subtitles \cite{perego-et-al-2010}. The reading speed shows how fast a user needs to read in order not to miss any part of the subtitle.
The lower the reading speed, the better is the model's output since a fast reading speed increases the cognitive load and leaves less time to look at the image.
The scrolling line method offers the best balance between latency and a comfortable reading speed
resulting
to be the best choice for the simultaneous scenario. 
On the other hand, this approach requires segmented text (i.e. a text that is divided into subtitles), thus 
the system needs to be able to simultaneously generate
transcripts or translations together with proper subtitle delimiters.
However, building a simultaneous subtitling system combines the difficulties of the simultaneous setting with the constraint of having a text formatted in proper subtitles. Since both these research directions are still evolving, a lot of work is required to achieve good results.

The lack of studies on this aspects highlights the shortcomings of the actual SimulST systems, individuating possible improvements that will allow the systems to evolve in a more organic and complete way according to the user needs.
Moreover, to completely assess the subtitling scenario, a system has to be able to jointly produce timestamps metadata linked to the word emitted, a task that has not been addressed so far. 
The need for this kind of system represents an interesting direction to follow for the simultaneous community. In the light of this, the researcher should also take into account the three quality-latency-visualization factors in their analyses.
We are convinced that these are the most promising 
aspects to work on to build the best SimulST system for the audience and that human evaluation has to have a crucial role in future studies.
We also believe that interdisciplinary dialogue with other fields such as cognitive studies, media accessibility and human-computer interaction would be very insightful to evaluate SimulST outputs from communicative perspectives \cite{fantinuoli-prandi-2021-towards}.

%

\section{Conclusions and Future directions}
SimulST systems have become increasingly popular in recent years and many efforts have been 
made
to build robust and efficient models.
Despite the difficulties introduced by the online framework, these models have rapidly improved, achieving comparable results to the offline 
systems.
However, many research directions have not been explored enough (e.g., the adoption of dynamic or fixed segmentation, the offline or the online training).
First among all, the visualization strategy that is adopted to display the output of the simultaneous systems is an important and largely under-analyzed aspect of the simultaneous experience. 
We posit that the presence of application-oriented metrics (e.g., reading speed), which are strongly related to the visualization and, as an ultimate goal, to the user experience, is the factor that misses in the actual evaluation environment.
Indeed, this
paper points out that BLEU and Average Lagging are not the only metrics that matter to effectively evaluate a SimulST model, even if they are fundamental to judge a correct and real-timed translation. We hope that this will inspire the community to work on this critical aspect in the future.

\section*{Acknowledgement}

This work has been carried out as part of the project Smarter Interpreting (\url{https://kunveno.digital/}) financed by CDTI Neotec funds.

\bibliographystyle{acl}
\bibliography{bibliography.bib}

\end{document}